\documentclass[sigconf]{aamas} 

\usepackage{subcaption}
\usepackage{wrapfig} 
\usepackage{float} 
\usepackage{multirow} 

\usepackage{balance} 

\doi{AJIK9102}

\makeatletter
\gdef\@copyrightpermission{
  \begin{minipage}{0.2\columnwidth}
   \href{https://creativecommons.org/licenses/by/4.0/}{\includegraphics[width=0.90\textwidth]{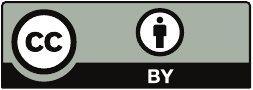}}
  \end{minipage}\hfill
  \begin{minipage}{0.8\columnwidth}
   \href{https://creativecommons.org/licenses/by/4.0/}{This work is licensed under a Creative Commons Attribution International 4.0 License.}
  \end{minipage}
  \vspace{5pt}
}
\makeatother

\setcopyright{ifaamas}
\acmConference[AAMAS '26]{Proc.\@ of the 25th International Conference
on Autonomous Agents and Multiagent Systems (AAMAS 2026)}{May 25 -- 29, 2026}
{Paphos, Cyprus}{C.~Amato, L.~Dennis, V.~Mascardi, J.~Thangarajah (eds.)}
\copyrightyear{2026}
\acmYear{2026}
\acmDOI{}
\acmPrice{}
\acmISBN{}

\acmSubmissionID{925}

\title[Modelling Customer Trajectories with Reinforcement Learning for Practical Retail Insights]{Modelling Customer Trajectories with Reinforcement Learning for Practical Retail Insights}

\author{Ken Ming Lee}
\affiliation{
  \institution{McGill University}
  \city{Montréal}
  \country{Canada}
}
\email{ken.m.lee@mail.mcgill.ca}

\author{Paul Barde}
\affiliation{
  \institution{Mila - Quebec AI Institute}
  \city{Montréal}
  \country{Canada}
}
\email{paul.b.barde@gmail.com}

\author{Maxime C.~Cohen}
\affiliation{
  \institution{McGill University}
  \city{Montréal}
  \country{Canada}
}
\email{maxime.cohen@mcgill.ca}

\author{Derek Nowrouzezahrai}
\affiliation{
  \institution{McGill University}
  \city{Montréal}
  \country{Canada}
}
\email{derek.nowrouzezahrai@mcgill.ca}

\begin{abstract}
Understanding customer movement within retail spaces is essential for optimizing store layouts. Real-world trajectory data can provide highly accurate insights, but collecting it is costly and often infeasible for many retailers. Heuristics such as Travelling Salesman Problem (TSP) and Probabilistic Nearest Neighbours (PNN) are commonly used as inexpensive approximations, but actual customer trajectories deviate by an average of 28\% from shortest paths, highlighting a tradeoff between accuracy and practicality. We propose an agent-based modelling framework that casts customer trajectory prediction as a maximum entropy reinforcement learning (RL) problem, balancing reward maximization with stochasticity to better reflect customers with bounded rationality. Using real-world trajectory data from a convenience store, we show that RL-generated trajectories align more closely with customer behaviour than TSP and PNN, providing more accurate estimates of impulse purchase rates and shelf traffic densities. Furthermore, only RL-based predictions yield repositioning decisions for impulse products that align with those derived from actual trajectory data, resulting in comparable estimated profit gains. Our work demonstrates that RL provides a practical, behaviourally grounded alternative that bridges the gap between oversimplified heuristics and data-intensive approaches, making accurate layout optimization more accessible. To encourage further research, the source code is available on GitHub.
\end{abstract}



\newcommand{\BibTeX}{\rm B\kern-.05em{\sc i\kern-.025em b}\kern-.08em\TeX}

\newcommand{\suppref}[2]{%
  \expandafter\newcommand\csname r@#1\endcsname{{#2}{0}}%
}

\suppref{appendix:rl-training-details}{A}
\suppref{appendix:tsp-pnn-implementation-details}{B}
\suppref{appendix:additional-heatmaps}{C}

\newcommand{\boldparagraph}[1]{\par\textbf{#1.}\hspace{.5em}}

\begin{document}


\pagestyle{fancy}
\fancyhead{}


\maketitle





\begin{figure*}
    \centering
    \begin{subfigure}[t]{.24\textwidth}
        \vspace{0pt}
        \includegraphics[width=\textwidth]{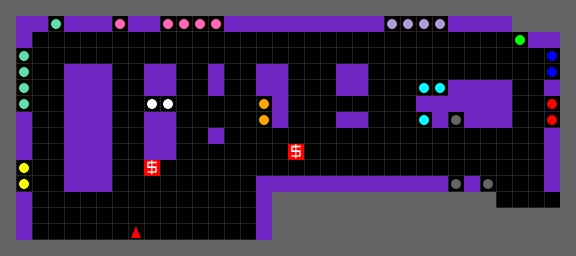}
         \caption{
         RL StoreGrid environment.
        }
    \end{subfigure}
    \hspace{.01\textwidth}
     \begin{subfigure}[t]{.24\textwidth}
        \vspace{0pt}
         \includegraphics[width=\textwidth]{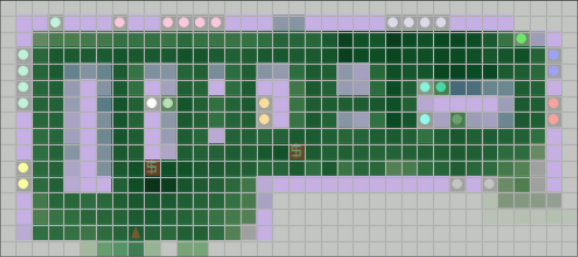}
         \caption{Overlay of customer trajectories on the RL environment.}
     \end{subfigure}
     \hspace{.01\textwidth}
     \begin{subfigure}[t]{.25\textwidth}
        \vspace{0pt}
         \includegraphics[width=\textwidth]{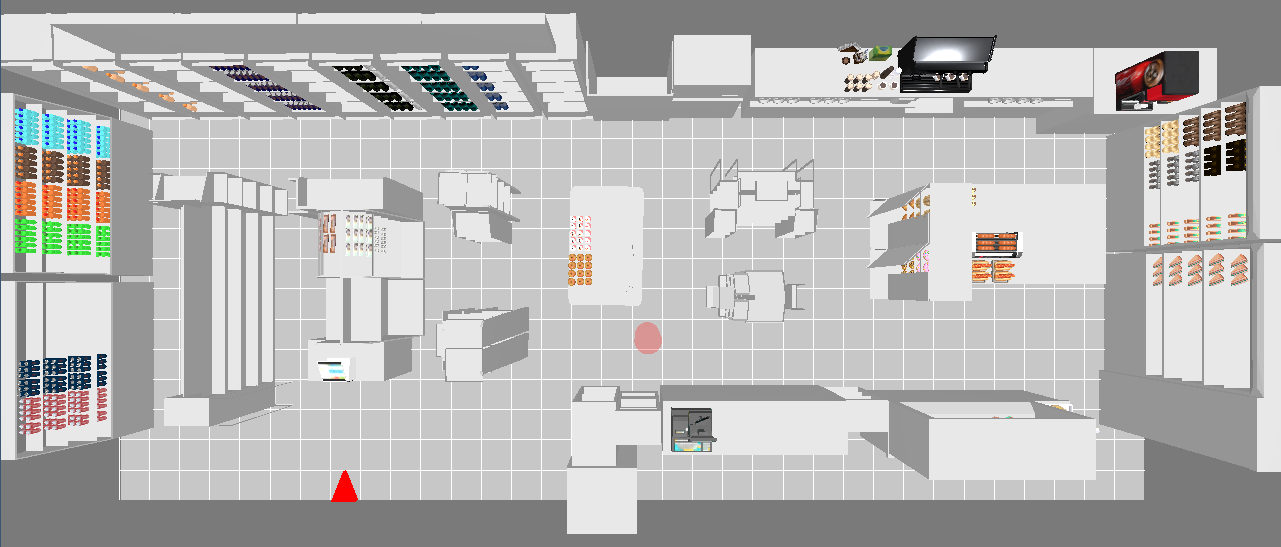}
         \caption{Top-down view of the digital twin.}
     \end{subfigure}
     \hspace{.01\textwidth}
     \begin{subfigure}[t]{.2\textwidth}
        \vspace{0pt}
         \includegraphics[width=\textwidth]{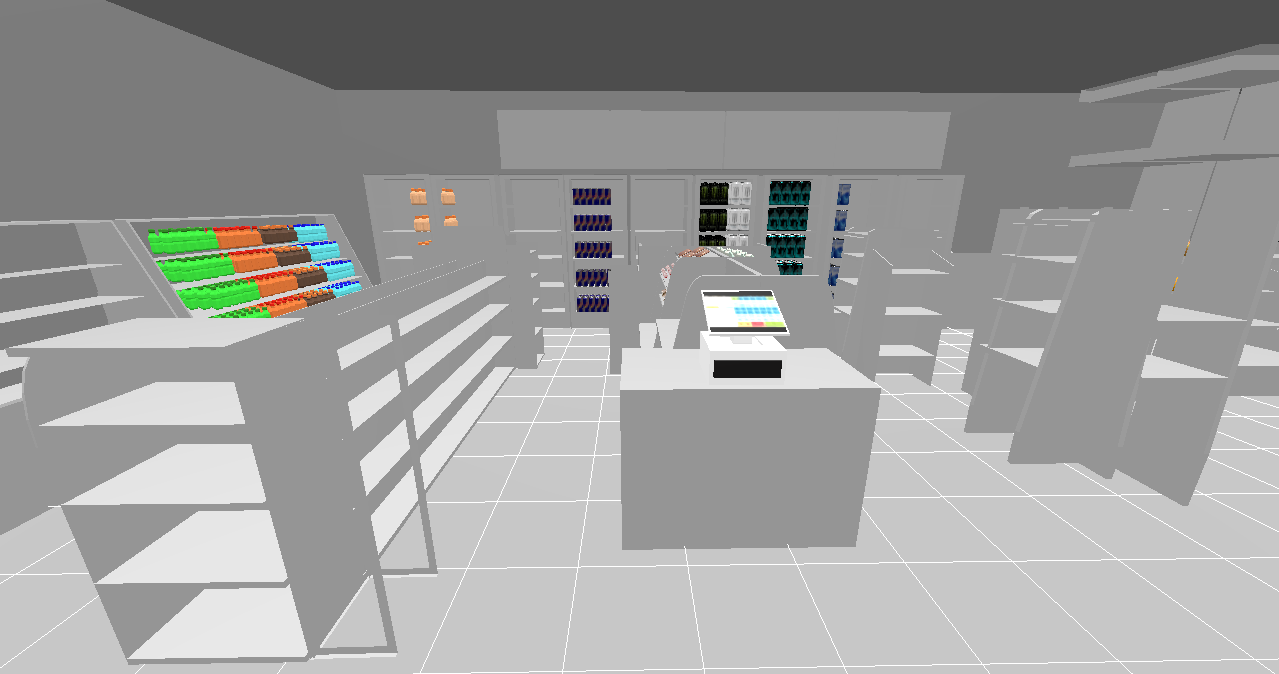}
         \caption{First-person view of the digital twin.}
     \end{subfigure}
     \Description{TODO}
     \caption{Various representations of the retail store. (a) Grid-based representation for RL training. (b) Overlay of customer trajectories on the RL environment, illustrating spatial alignment with the physical store. (c-d) Top-down and first-person views of the 3D digital twin constructed from the same discretized representation.}
\label{fig:map}
\end{figure*}
\section{Introduction}
Maximizing profit is the central goal of store layout optimization.
This is typically achieved by strategically arranging products to maximize impulse purchases, which, unlike planned purchases, are highly influenced by store layout and customer exposure \cite{impulse1995Rook}.
Therefore, understanding customer movement is paramount in determining the optimal placement for each product.

While customer trajectories can provide these insights, collecting them is costly, making it infeasible for many retailers, particularly smaller ones.
As an alternative, heuristics such as the Travelling Salesman Problem (TSP) and Probabilistic Nearest Neighbour (PNN) are commonly used in the literature to approximate customer paths \cite{flamand-2023, hirpara2021retail, dorismond2023simulation, holmgren2021customer, ozgormus2020data}. 
TSP assumes the global shortest path, while PNN models a stochastic greedy-optimal route.
However, prior research shows that customers deviate from shortest paths by an average of 28\% \cite{hui2009research}, highlighting a gap resulting from a tradeoff between these approaches: heuristics provide inexpensive but unrealistic approximations, while actual trajectories offer accuracy but at prohibitive cost.

To address this gap, we propose an agent-based modelling framework in which customers are assumed to follow trajectories derived from maximum entropy reinforcement learning (RL).
Maximum entropy RL optimizes for both reward and randomness simultaneously \cite{ziebart2008maximum}, making it a compelling alternative for modelling customers with bounded rationality. 
Using ground-truth trajectory data collected from a convenience store, we demonstrate that RL-generated paths more closely resemble real-world customer behaviour than those generated by TSP and PNN, as measured by several divergence metrics.
These more realistic trajectories translate to more accurate estimates of impulse purchase rates and shelf traffic densities. 
Furthermore, when using trajectories generated by these methods to inform the repositioning of a single impulse product, only RL-based predictions yield decisions consistent with those derived from actual customer trajectories. 
Simulating customer trajectories on these revised layouts to estimate profit outcomes further confirms this advantage: only the RL-informed layout achieves a profit increase comparable to that obtained using ground-truth trajectory data.

These findings show that RL-generated trajectories provide a more accurate approximation of customer behaviour than heuristic methods, while serving as a cheaper substitute for actual trajectories. 
In doing so, RL bridges the gap between oversimplified heuristics and data-intensive approaches, making accurate layout optimization more accessible.

The supplementary material, the link to the source code, and a playable demo of the digital twin are publicly available at \url{https://sites.google.com/view/storegrid}.

\section{Background Information} \label{chapter:background-info}
The facility layout problem studies the optimal arrangement of physical spaces (e.g., workstations, shelves) to achieve operational goals. \cite{al2021analysis}.
In manufacturing, a typical objective is to minimize material handling costs by optimizing the flow of goods between production units \cite{drira2007facility}. 
In contrast, the retail setting instead prioritize profitability and customer experience \cite{botsali2007retail}, making behavioural modelling essential.
Figure~\ref{fig:map} illustrates the retail store layout used throughout this paper.

\subsection{Retail Layout Design Subproblems}
Retail layout design is a hierarchical process spanning aisle configuration (e.g., grid, freeform, racetrack layouts)\cite{li2011facility}, zoning (i.e., assigning departments to functional zones) \cite{dorismond2023simulation, danisman2023data}, shelf-space allocation (i.e., how much shelf space to allocate to each product or category) \cite{ghoniem2016optimization, flamand2016promoting}, and product placement (i.e., optimal physical arrangement of items) \cite{edirisinghe2023strategic, corstjens1981model, corstjens1983dynamic}.

This paper is primarily concerned with the product-placement subproblem. 
Accordingly, all subsequent references to "store layout optimization" refer specifically to this component of the broader retail layout design process.

\subsection{Impulse Purchases}
In the retail literature, customer purchases are typically categorized into planned and unplanned purchases.
Planned purchases refer to items that customers intend to buy before entering the store (e.g., essential products like bread and milk), and are largely unaffected by layout.
In contrast, impulse purchases are spontaneous and heavily influenced by in-store stimuli such as product placement, displays, or advertisements\cite{impulse1995Rook}. 
Impulse purchases can account for 30–50\% or more of total sales in supermarkets \cite{kollat1967customer, popai2014, bellenger1978impulse, underhill2009we}, making them the primary source of marginal profit gain from layout optimization. 

Hence, much of the literature on product-level layout optimization focuses specifically on maximizing impulse profit — the portion of profit attributable to unplanned purchases influenced by the in-store environment.

\subsection{Optimizing Store Layouts to Maximize Impulse Purchases}
A central principle in the retail literature is "unseen is unsold" \cite{gul2023retail, hui2013effect}, motivating layouts that maximize product exposure. 
Earlier strategies attempt to increase exposure by distributing essential products throughout the store to lengthen paths \cite{granbois1968improving, iyer1989unplanned, bhadury2016store}, but a limitation with this approach is that excessive detours risk reducing shopping convenience and thus overall sales \cite{botsali2023effect, juel2015aisles}. 

To address this limitation, subsequent studies have sought to balance between maximizing exposure and customers' convenience. 
For instance, \citet{ozgormus2020data} alternates between optimizing for revenue and adjacency preferences, while \citet{abdelaziz2024store} formulates the problem as a bi-objective constraint optimization problem, where the authors maximizes impulse profit while constraining shopping distance of customers.

\subsection{Evaluating Store Layouts}
Effectively addressing the product-placement subproblem requires not only a strong optimization algorithm but also an evaluation framework that closely aligns with the ultimate goal of profit maximization.
In the literature, evaluation of layouts is typically either integrated directly into the optimization objective — as in constraint optimization approaches (e.g., \citet{flamand-2023}) — or performed explicitly as a separate step \cite{dorismond2016, dorismond2023simulation, holmgren2021customer}.

Regardless of the approach, most evaluation frameworks rely on predicting customer navigation patterns, where  heuristics such as the Travelling Salesman Problem (TSP) and Probabilistic Nearest Neighbour (PNN) dominate \cite{flamand-2023, hirpara2021retail, dorismond2023simulation, holmgren2021customer, ozgormus2020data}. 
TSP assumes globally shortest routes, while PNN stochastically select the next item based on proximity, assigning probabilities inversely proportional to distance.

Although inexpensive, prior research shows that actual customer trajectories can deviate from shortest-path estimates by an average of 28\% \cite{hui2009research}, highlighting the limitations of such models in capturing realistic customer behaviour.
This creates a gap between costly but accurate trajectory data and oversimplified heuristic approximations.

\subsection{Maximum Entropy Reinforcement Learning}
In Reinforcement Learning (RL), an agent learns a policy, that selects an action given a state, to maximize expected cumulative reward through sequential interaction with an environment \cite{sutton2018reinforcement}. 
Maximum Entropy (MaxEnt) RL augments this by jointly maximizing the expected return and the policy entropy \cite{ziebart2008maximum}. 
This encourages diverse behaviours and captures multi-modal solutions \cite{haarnoja2017reinforcement, haarnoja2018soft}. 

In the case of retail, for a given shopping basket, customers may not only  purchase products in different orders, but also take different paths for a specific product sequence.
Traditional heuristics like TSP and PNN are unable to capture such behavioural variability.
Moreover, the space of all possible trajectories through the store grows combinatorially with the number of products to pickup and size of the store, quickly becoming too large for exhaustive enumeration.  

In contrast, since maximizing randomness is part of the objective of a MaxEnt RL algorithm, it is naturally incentivized to explore diverse trajectories, including different product-pickup orders or route variations.
Additionally, when used in combination with expressive function approximators such as neural networks, MaxEnt RL can scale to explore large trajectory spaces much more effectively than heuristic or exact methods.

\section{Experimental Setup} \label{chapter:experimental-setup}
This section describes the experimental setup, RL formulation of the physical store, preprocessing of real-world trajectory data, and implementation of trajectory modelling methods.

\subsection{Modelling a Real-World Convenience Store}
We collaborated with a local convenience store outfitted with overhead cameras arrays, and obtained anonymized customer trajectory data collected from September 2023 to February 2024. 
The dataset contained 3D joint coordinates recorded at 5 Hz and corresponding checkout baskets.
From this, we reconstructed customers' 2D in-store positions and associated purchased items.
Layout metadata describing boundaries of shelves and checkout locations were obtained from \citet{li2023dynamic}.

To compare trajectory-prediction methods, we implemented a 2D gridworld environment (Figure~\ref{fig:map}) based on the Gymnasium Minigrid framework \cite{MinigridMiniworld23}.
The environment matched the store’s actual dimensions: a $16\times36$ grid with each cell representing $50\times50$ cm. 
This discretization allows the store to be represented as a graph for pathfinding (e.g., Dijkstra’s algorithm) and defines a finite action space for RL agents, improving computational tractability.

We focused on the top 61 best-selling products, which account for approximately 51\% of sales.
Products were grouped into 11 categories: Hot Coffee/Tea, Bakery/Pastries, Hot Food (e.g., Hot Dog, Pizza), Fruits/Yogurt, Energy Drinks, Cold Beverages (e.g., Kombucha, Sparkling), Soft Drinks (e.g., Coke, Pepsi), Snack Bars (e.g., Energy, Granola), Cold Food (e.g., Sandwiches, Wraps, Salad), Cold Coffee/Tea/Shake and Fountain Drinks. 

Each shelf holds one category, and both self-checkout and cashier-assisted stations are represented as distinct checkout locations (shown as red cells with dollar signs in Figure~\ref{fig:map}).
For convenience, the term "product" refers to the entire category throughout this paper.

\subsection{Preprocessing Trajectory Data}
Raw trajectories are preprocessed to align with the gridworld environment as follows.
Trajectories without matching basket information are removed, and coordinates outside store boundaries or recorded after checkout are trimmed.
Continuous $(x, y)$ positions are then discretized to grid indices, and invalid points (e.g., within shelves or walls) are reassigned to the nearest valid cell.
All trajectories are normalized to start at the store entrance and end at one of the two checkout counters.
Finally, each trajectory is transformed into a sequence of $(state, action)$ pairs, with pickup actions inferred at the customer’s nearest (or latest) approach to a product location.
After preprocessing, 3{,}054 trajectories remained for evaluation purposes.

\subsection{RL Environment and Agent Design}
We model customers as conditional MaxEnt RL agents, trained using Proximal Policy Optimization (PPO) \cite{schulman2017proximal} with a convolutional neural network backbone.

Each episode represents a complete shopping trip: the agent starts at the entrance and ends upon performing a checkout action or reaching a time limit.
The agent is conditioned on the target basket (i.e., products to purchase), checkout location (two exist in this case), and optionally, a timestep budget (i.e., target trip length).
Conditioning on timestep budget enables control over the length of generated trajectories, allowing us to simulate customers with different shopping preferences.

The agent observes a multi-layer 2D tensor that encodes environmental features.
Specifically, the observation includes:
\begin{enumerate}
    \item an object-type map (indicating walls, shelves, products etc.);
    \item step count in the current episode;
    \item timestep budget (if desired);
    \item a binary mask marking cells yet to be visited;
    \item category identifiers for each product;
    \item conditional basket specifying the customer’s intended purchases; and
    \item the agent’s current position and orientation.
\end{enumerate}
The discrete action space comprises four actions: move forward, turn left, turn right, and pickup/checkout.

At checkout, the agent receives a reward based on the accuracy of products picked-up, use of the correct checkout point, and adherence to the specified timestep budget. Rewards scale with the proportion of correctly collected items and closeness to the target trip length.

Additional details, including RL network architecture and hyperparameters settings, are provided in Appendix \ref{appendix:rl-training-details}.

\subsection{Training Techniques}
To improve learning stability and promote generalizable policies, several practical techniques are applied:
\begin{itemize}
\item Conditional baskets, ranging from 0–5 products (which are broader than the real product range), are periodically resampled to improve generalization across basket sizes and product combinations. 
Training follows a curriculum schedule, progressing from simpler to more complex baskets.
\item Parallel environments with independently sampled basket conditions are used to stabilize learning and enhance generalization.
\item Observations are normalized per channel to account for differing value ranges.
\item A discount factor of $\gamma = 1.0$ ensures that longer trajectories are not penalized, preserving natural path-length variability and supporting time-conditioned control.
\item To further promote exploratory behaviour, a bonus reward is granted after all main objectives have been satisfied, based on the number of unique states visited in an episode.
\end{itemize}

\subsection{Trajectory Generation}
We generate TSP and PNN trajectories by treating the RL environment as a graph, with adjacent cells as connected nodes. 

For the TSP baseline, the shortest possible route is computed, whereas for the PNN baseline, the next product is selected probabilistically, with nearer products more likely to be chosen. In both cases, the checkout is appended as the final waypoint. Detailed description of their implementations are available in Appendix \ref{appendix:tsp-pnn-implementation-details}.

To generate RL trajectories, the trained RL policy is conditioned on a basket, a checkout location, and an optional timestep budget -- all of which are encoded into the agent’s observation.
We then roll out the policy in the environment and only retain trajectories that exceeds a minimum reward threshold, ensuring the specified conditions are satisfied.

\section{Experimental Results} \label{chapter:experimental-results}
We evaluate our agent-based modelling framework through three stages:
(1) comparing how closely different trajectory-generation methods reproduce real human movement,
(2) assessing their ability to estimate shelf-level visitation and impulse rates, and
(3) applying these estimates to reposition an impulse product for profit.

In this section, we describe these stages in detail, and present results alongside a corresponding discussion of them.
\subsection{Performance Comparison of Trajectory Generation Methods} \label{chapter:performance-comparison-between}
We sample 10k trajectories for each method across  all baskets combination present in the ground-truth dataset.
For those with fewer than 10k available trajectories (such as TSP and human trajectories), we upsample with replacement to ensure fair comparison.

To evaluate each method's ability to replicate real-world customer behaviour, we aggregate their trajectories for each basket and normalize them into 2D probability distributions.
They are compared to the real human trajectory distribution using Jensen-Shannon divergence (JSD) and Wasserstein Distance (WD), where smaller values indicate closer alignment with human behaviour.

From Table~\ref{table:avg-heatmap}, it can first be observed that PNN's stochastic sampling of waypoints across all possible product locations allows it to better capture the diversity and multimodality present in real-world customer behaviour, 
allowing it to outperform TSP in terms of JSD and WD.
\begin{figure*}
    \centering
    \includegraphics[width=.9\textwidth]{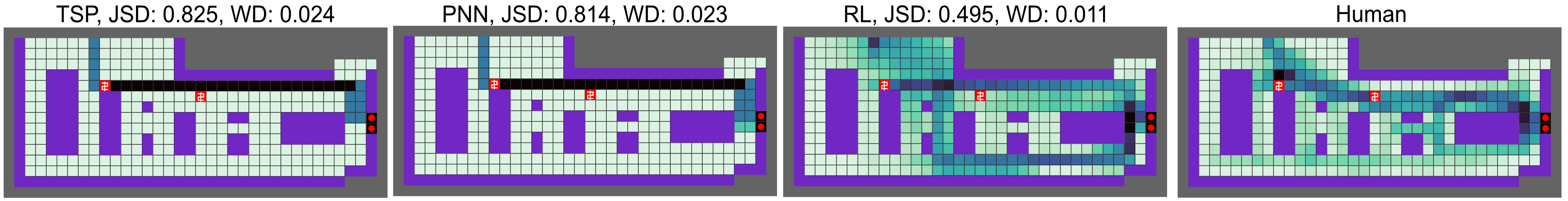}
    \caption[Customer trajectory heatmaps for Cold Food (checkout 9,5)]{Trajectory heatmap of customers purchasing from the Cold Food category with (9,5) checkout, across all methods. 
    Only RL was able to recover the mode where the customer purchases the product by travelling the longer way from the bottom.}
    \label{fig:rl_success_mode_(9,5)}
    \Description{TODO}
\end{figure*}
\begin{table}
    \centering
     \caption[Average divergence metrics (JSD, WD) comparing methods]{Divergence of simulated trajectories compared to human trajectories. 
     Lower (bolded) scores indicate closer alignment with human behaviours.}
     \label{table:avg-heatmap}

    \begin{tabular}{llll}
        \toprule
        Divergence (lower is better) & TSP & PNN & RL \\ 
        \midrule
        Jensen-Shannon divergence (JSD) & 0.657 & 0.580 & \textbf{0.415} \\
        of average heatmap & & \\
        Wasserstein Distance (WD) of & 0.0140 & 0.0120 & \textbf{0.00800} \\
        average heatmap & & \\
        Average JSD & 0.777 & 0.676 & \textbf{0.476} \\
        Average WD & 0.0176 & 0.0142 & \textbf{0.00920} \\
        \bottomrule
    \end{tabular}
\end{table}

However, PNN falls short compared to RL.
Figure \ref{fig:rl_success_mode_(9,5)} illustrates this qualitatively:
TSP fails to capture the mode where the customer navigates through the bottom half of the store to reach the product, because it always selects the shortest global path, which in this case, leads through the top of the store. 
Although PNN is stochastic in choosing which shelf to visit next, the shortest path to both shelf locations of the Cold Food category happens to go through the upper half of the store, causing PNN to similarly miss the bottom traversal.
In contrast, since the RL agent is trained to maximize not only reward but also policy entropy, it is incentivized to explore and recover diverse yet task-valid trajectories, such as the one through the bottom half of the store. 

In summary, RL trajectories align more closely with real-world customer behaviour than those generated by either PNN or TSP. 
This trend is consistent across both JSD and WD, whether computed over average occupancy heatmaps (Table \ref{table:avg-heatmap}) or examined on a per-basket basis (Appendix \ref{appendix:additional-heatmaps}).

\subsection{Use Case 1: Estimating Traffic Density}
To maximize impulse purchases, high-impulse items should be placed in areas with high foot traffic. 
In contrast, essential items — which are purchased regardless of placement — can be used strategically to guide customers through high-value zones \cite{holmstrom1997product}. 
To take advantage of this, it's critical to quantify how likely each shelf is to be visited during an average customer trip.

For instance, \citet{flamand-2023} proposed a profit-maximizing store layout formulated as a constrained optimization problem, with the objective:

\begin{equation}
    J_{\text{profit}} = \max_x \sum_{b\in \mathcal{B}} \sum_{p \in \mathcal{P}} \rho_p\, i_p\, \frac{s_{pb}}{c_b}\, x_{bp}\, \theta_b
    \label{eqn:impulse-rate-profit}
\end{equation}

where 
\begin{itemize}
    \item $b$ indexes shelves, and $p$ indexes product categories,
    \item $\rho_p$ denotes the per-unit profit of product category $p$, which can be expressed as $\rho_p = \text{price}_p \times \text{margin}_p$,
   \item $i_p$ represents impulse rate of product category $p$,
   \item $\frac{s_{pb}}{c_b}$ is the ratio between space taken up by product $p$ and total shelf space $c_b$; conceptually, this ratio measures visibility of $p$ on shelf $b$, which equals 1 in our case (one category per shelf),
   \item $x_{bp}$ is a binary placement variable to optimize for, and
   \item $\theta_b$ is the \emph{shelf traffic density} — the probability that shelf $b$ is visited during a typical store visit.
\end{itemize}

Intuitively, the shelf traffic density $\theta_b$ depends not only on the shelf's physical location in the store, but also on the shelf's content. 
The closer the shelf is to the entrance/exit, the more likely it is to be visited. 
Likewise, if the product placed on the shelf is popular, or the shelf is close to another popular product, the shelf is more likely to be visited as well.
To predict the traffic density of a shelf in a new layout, \citet{flamand-2023} train a regressor to do so, which takes it's and neighbouring shelves' contents into consideration.

Importantly, since \citet{flamand-2023} did not have access to human trajectory data, they instead used simulated TSP trajectories as the ground-truth dataset for training the regressor. 
Motivated by previous results where RL-generated trajectories more closely resemble real customer movement, we propose deriving shelf traffic density from RL-generated trajectories instead of TSP (and PNN).

\begin{figure}
    \centering
    \begin{subfigure}[t]{.49\linewidth}
         \includegraphics[width=\linewidth]{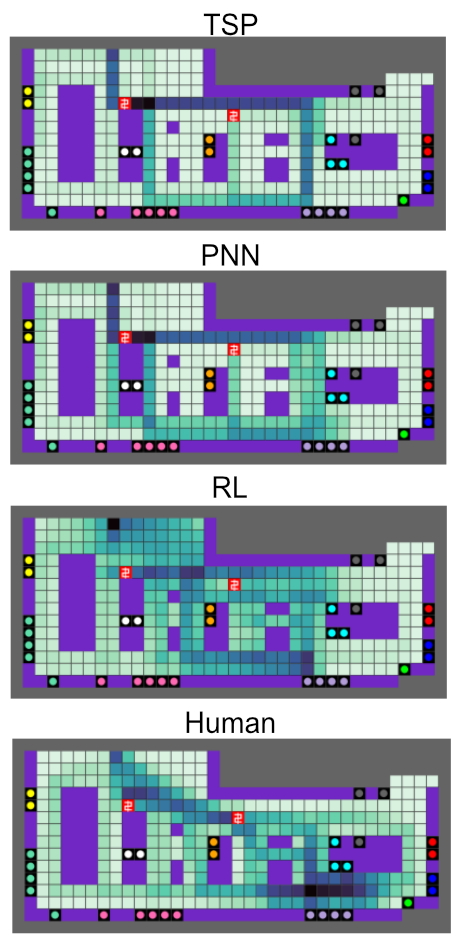}
         \caption{Trajectory heatmap. }
         \label{fig:usecase1-heatmap}
     \end{subfigure}
    \begin{subfigure}[t]{.49\linewidth}
         \includegraphics[width=\linewidth]{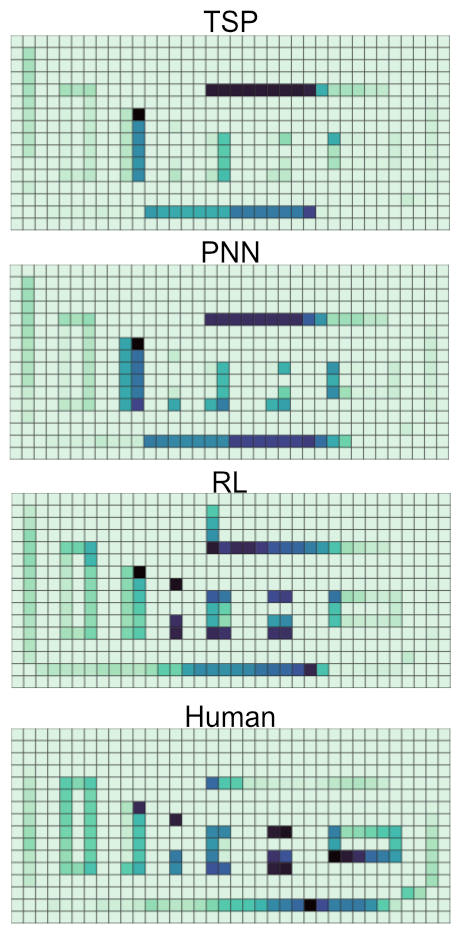}
         \caption{Heatmap of shelf-traffic density computed from Figure~\ref{fig:usecase1-heatmap}.}
         \label{fig:usecase1-shelves-visitation-proba}
     \end{subfigure}
    \caption[Aggregated trajectory and shelf-traffic heatmaps for top 61 baskets]{Heatmap of trajectories and shelf-traffic density generated by TSP, PNN, RL and ground-truth (human) data, for the top 61 most frequently bought baskets.}
    \label{fig:usecase1}
    \Description{TODO}
\end{figure}

To compute shelf traffic density from trajectories, for each trajectory, we iterate through each shelf in the store, and mark a shelf as being visited if the customer gets close to the shelf (where "close" is defined as reaching an adjacent cell).
The resulting shelf occupancy map is then normalized by the number of trajectories, yielding a 2D grid where each value represents the average probability that a shelf at that location is visited during a trip (see Figure~\ref{fig:usecase1-shelves-visitation-proba}).

To evaluate the quality of the shelf traffic density prediction by each method, we compute their JSD and WD to that of ground truth human trajectories. 
From Table \ref{table:usecase1}, it can be observed that RL yields lower JSD and WD than TSP and PNN.
This is likely attributed to TSP and PNN's limited coverage over certain shelves, especially for shelves that may not be passed by using shortest path algorithms (see Figure~\ref{fig:usecase1-shelves-visitation-proba}).

\begin{table}
    \centering
     \caption[Comparison of shelf traffic divergence under different sampling schemes]{Divergence between shelf traffic density of different methods and that of ground-truth human trajectories. 
     Lower (bolded) scores indicate closer alignment with human behaviour.}
     \label{table:usecase1}
     
    \begin{subtable}[t]{\linewidth}
        \centering
       \caption{\textbf{Proportional Sampling}: Basket trajectories sampled proportionally to real-world purchase frequencies, potentially biasing results toward popular baskets.}
        \label{table:shelves-traffic-density-proportionate-sampling}
        \begin{tabular}{llll}
            \toprule
            Divergence & TSP & PNN & RL \\ 
            \midrule
            JSD & 0.632 & 0.549 & \textbf{0.430} \\
            WD & 0.313 & 0.278 & \textbf{0.217} \\
            \bottomrule
        \end{tabular}
    \end{subtable}
    
    \vspace{1em}
    
    \begin{subtable}[t]{\linewidth}
        \centering
       \caption{\textbf{Uniform Sampling}: Each basket sampled equally, avoiding over-representation of popular baskets. 
       All divergences reported are measured against uniformly-sampled human trajectories. 
       The last column shows divergence between proportionally and uniformly sampled human trajectories to illustrate effect of sampling choice. }
        \label{table:shelves-traffic-density-uniform-sampling}
        \begin{tabular}{llll|l}
            \toprule
            Divergence & TSP & PNN & RL & Human\\
            \midrule
            JSD & 0.505 & 0.506 & \textbf{0.347} & 0.224 \\
            WD & 0.0106 & 0.0106 & \textbf{0.00676} & 0.00517 \\
            \bottomrule
        \end{tabular}
    \end{subtable}
\end{table}

\subsection{Use Case 2: Estimating Impulse Rates} \label{section:usecase2}
To optimize store layouts for impulse profit, it is insufficient to only know which shelves receive the most traffic; we must also estimate which products are most likely to be purchased spontaneously — that is, which products exhibit the highest impulse rates.

Prior approaches have estimated impulse rates using domain expertise \cite{ozgormus2020data}, customer surveys \cite{azar2023store}, or by assuming a direct correlation with purchase probabilities \cite{dorismond2023simulation}. 
However, domain knowledge may be unreliable or unavailable, especially for new or smaller retailers.
Surveys are resource-intensive to conduct and may not scale well.
Purchase probability is not a reliable proxy for impulse rate, since essential products are frequently purchased but rarely on impulse.

In our case, access to simulated customer trajectories enables a data-driven estimation of impulse rates using the following formulation adapted from \citet{azar2023store}:
\begin{equation}\label{eqn:purchase-proba}
   P_{\text{purchase}} = P_{\text{visit\_shelf}} \times P_{\text{prod\_visibility}} \times P_{\text{impulse rate}} 
\end{equation}
where for a given product,
\begin{itemize}
    \item $P_{\text{purchase}}$ is the probability that the product is purchased,
    \item $P_{\text{visit\_shelf}}$ represents the probability that the product's corresponding shelves are visited,
    \item $P_{\text{prod\_visibility}}$ is the probability that the product is visible to the customer upon visiting the shelf, and
    \item $P_{\text{impulse rate}}$ is the probability that the product is spontaneously purchased upon being seen.
\end{itemize}

In our case, each shelf contains only a single product category, so $P_{\text{prod\_visibility}}=1$,  
simplifying Equation~\ref{eqn:purchase-proba} to:
\begin{equation}
    P_{\text{purchase}} = P_{\text{visit\_shelf}} \times P_{\text{impulse rate}} 
    \label{eqn:visitation-proba}
\end{equation}

This equation expresses the intuition that a product can only be purchased if the shelf is visited (and the product seen), and then selected impulsively.

\begin{table}
    \centering    
   \caption[Purchase probabilities of product categories across customer clusters ]{
    Purchase probabilities ($P_{\text{purchase}}$) for each product category across three customer clusters (Cluster 1-3). 
    }
    \label{table:cluster-basket-probs}
    \begin{tabular}{llll}
        \toprule
        Product Category & Cluster 1 &  Cluster 2 &  Cluster 3 \\
        \midrule
        Hot Coffee/Tea & 0 & 0.642 & 1 \\ 
        Bakery/Pastries & 0 & 1 & 0 \\ 
        Hot Food  & 0.145 & 0 & 0 \\ 
        Fruits/Yogurt & 0.0427 & 0.0363 & 0 \\ 
        Energy Drinks & 0.287  & 0 & 0 \\ 
        Cold Beverages& 0.274 & 0 & 0.00809 \\ 
        Soft Drinks& 0.0668 & 0.0130 & 0 \\ 
        Snack Bars & 0.0438 & 0 & 0 \\ 
        Cold Food & 0.0296 & 0 & 0 \\ 
        Cold Coffee/Tea/Shake & 0.0482 & 0 & 0 \\ 
        Fountain Drinks & 0.0635 & 0 & 0 \\ 
        \bottomrule
    \end{tabular}
\end{table}

\begin{wrapfigure}{l}{.43\linewidth}
\includegraphics[width=\linewidth]{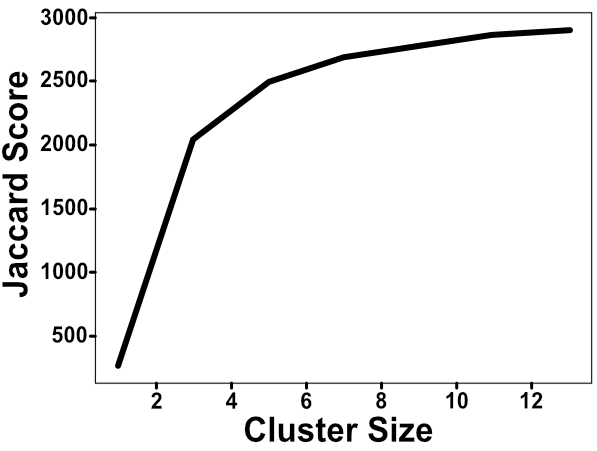} 
\caption{Within-Cluster Sum of Squares score against the number of clusters.}
\label{fig:wcss-plot}
\end{wrapfigure}

To compute $P_{\text{purchase}}$, we follow the method used by \citet{dorismond2023simulation} and clustered all 61 baskets using the elbow method, resulting in three clusters (Figure~\ref{fig:wcss-plot}).  
Computing the average basket per cluster then provides us with $P_{\text{purchase}}$ values for each product category (Table \ref{table:cluster-basket-probs}). 
Following their definition, within each cluster, products with $P_\text{purchase}<20\%$ are classified as impulse products.

To estimate $P_{\text{visit\_shelf}}$, we simulate customers purchasing solely essential (i.e., non-impulse) products.
This reflects the nature of impulse products: they should be picked up spontaneously, not as part of a preplanned list. 
For products stored across multiple shelves, we compute $P_{\text{visit\_shelf}}$ as the sum of visit probabilities across all relevant shelves (capped to a maximum of 1). $P_{\text{impulse rate}}$ is then computed by dividing $P_{\text{purchase}}$ by $P_{\text{visit\_shelf}}$ for each impulse product, per Equation \ref{eqn:visitation-proba}.

\begin{table}
    \centering
    \caption[Estimated impulse purchase rates for Cluster 2 customers]{
    Estimated impulse purchase rates across all methods for Cluster 2's impulse products . 
    \texttt{Inf} indicates a product was purchased despite no recorded shelf visits.
    }
    \label{table:usecase2-impulse-rates}
    \begin{tabular}{llll | l}
         \toprule
        Product Category & PNN & TSP & RL & Human\\
        \midrule
        Fruits/Yogurt & \texttt{Inf} & \texttt{Inf} & 0.0577 & 0.115 \\
        Soft Drinks& \texttt{Inf} & \texttt{Inf} & 7.20 & 3.20 \\
        \bottomrule
    \end{tabular}
\end{table}

As a concrete example, consider Cluster 2 in Table \ref{table:cluster-basket-probs}, where Hot Coffee/Tea and Bakery/Pastries are essential items, and Soft Drinks and Fruits/Yogurt are impulse products. 
10,000 trajectories are generated, such that customers purchase Hot Coffee/Tea with a probability of 0.642, and Bakery/Pastries with a probability of 1. 
Checkout destinations are sampled to match the empirical checkout distribution observed in the original dataset.

\begin{figure}
    \centering
    \includegraphics[width=\linewidth]{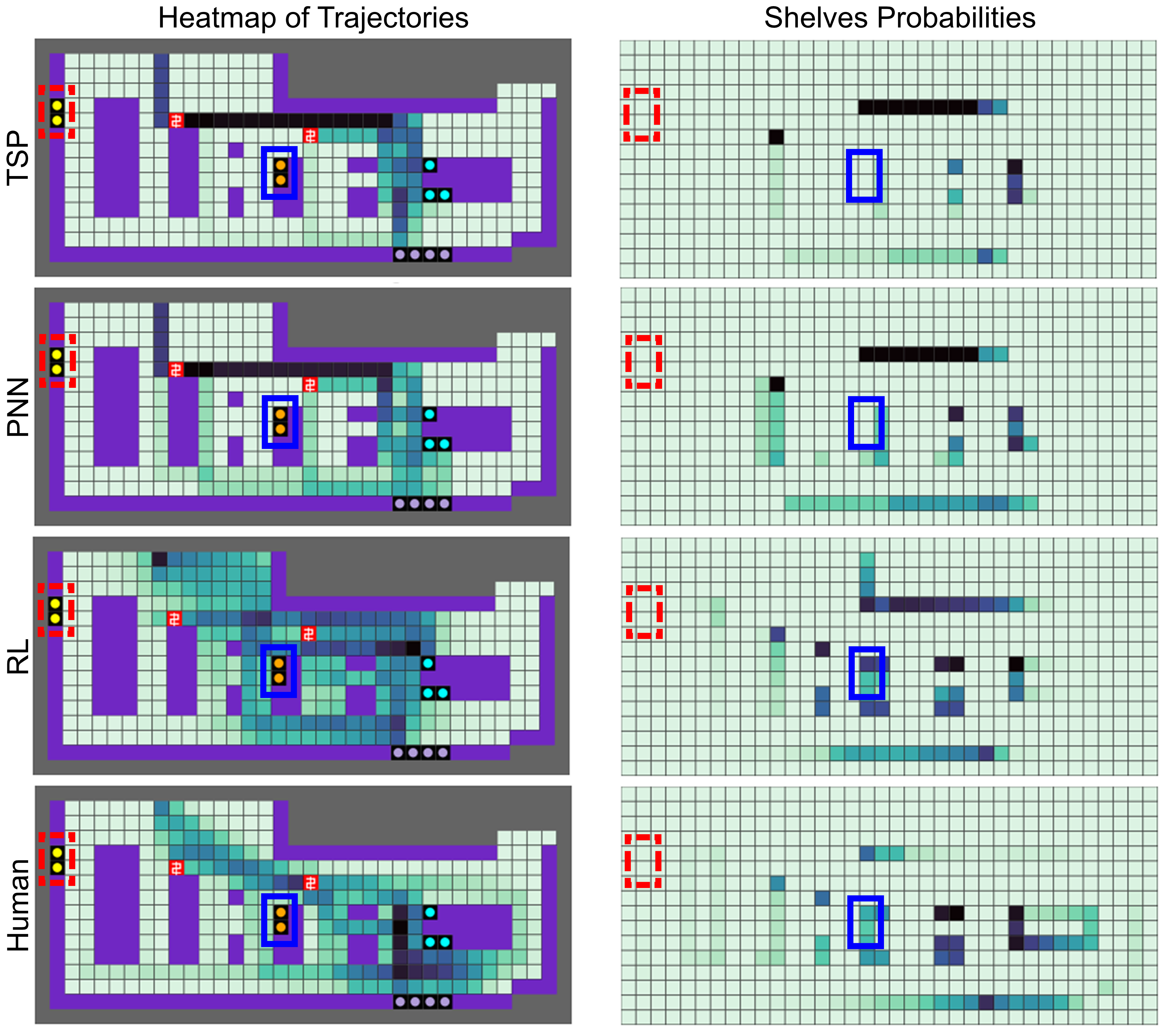}
    \caption[Trajectory heatmaps and shelf traffic density of Cluster 2 customers]{Left column shows trajectory heatmaps for Cluster 2; right column shows the corresponding shelf traffic density.
    Red (dotted) and blue (solid) boxes denote shelf regions for Soft Drinks and Fruits/Yogurt, respectively. 
    It can be observed that generated trajectories for TSP and PNN have not visited these shelves.}
    \label{fig:usecase2}
    \Description{TODO}
\end{figure}

From Table~\ref{table:usecase2-impulse-rates}, it can be observed that the generated trajectories of TSP and PNN miss the shelves housing both impulse items entirely, resulting in a division by zero when computing impulse rates via Equation \ref{eqn:visitation-proba}. 
This is likely because the shelves of Cluster 2's impulse products (i.e., Soft Drinks and Fruits/Yogurt) lie outside the direct routes connecting its essential products to checkout counters (illustrated at Figure~\ref{fig:usecase2}), hence posing a challenge for shortest-path based methods. 

In contrast, RL-generated trajectories exhibit more diverse routing behaviour and successfully reach the shelves containing Cluster 2's impulse items.
As observed in Table \ref{table:usecase2-impulse-rates}, while the estimated absolute values deviate somewhat from the ground truth, the relative ordering is preserved — Soft Drinks appear more impulsive than Fruits/Yogurt. 
This insight is particularly valuable for downstream product placement /promotional tasks.

\subsection{Use Case 3: Informing Layout Changes} \label{chapter:usecase3}
Having computed both impulse rates and shelf traffic densities, we now demonstrate how these metrics can inform concrete layout decisions to improve store profitability. 
For instance, these values can be directly used in optimization formulations such as those proposed by \citet{abdelaziz2024store} and \citet{flamand-2023}.
In our case, we utilize these findings to reposition a single impulse product to maximize store profit, demonstrating tangible improvements that these insights can bring to the store.

\boldparagraph{Step 1: Identifying the Most Profitable Impulse Product}
We begin by selecting the impulse product with the highest potential for profit. 
This is determined using the following heuristic, which estimates the expected impulse profit for product $p$. 
It is derived from the profit maximization objective in Equation~\ref{eqn:impulse-rate-profit}, under the assumption that all shelves (and thus all products) are always visited:
\begin{align}
   \pi_p &= i_{p} \times \rho_p \nonumber \\
         &= i_{p} \times \text{price}_p \times \text{margin}_p 
    \label{eqn:single-product-impulse-profit}
\end{align}
Here, $\pi_p$ denotes the expected impulse profit of product $p$, $i_p$ is the impulse rate, and $\rho_p$ is the per-unit profit (price times margin).
Due to the lack of product-level profit margin data, we assume a uniform margin of 5\% based on industry estimates \cite{sharpsheetsProfitableConvenience}.

Consistent with Use Case 2, we focus on Cluster 2, where Soft Drinks and Fruits/Yogurt are impulse products. 
Since TSP and PNN trajectories do not intersect with the shelves of both of these products, their impulse rates $i_p$ are approximated using purchase probabilities (from Table~\ref{table:cluster-basket-probs}) instead, with the assumption that higher purchase frequency correlates with higher impulsivity.

Table \ref{table:usecase3-estimated-profit} compares the estimated impulse rates and per-unit profits across all methods.
While TSP and PNN rank Fruits/Yogurt as more profitable, RL identifies Soft Drinks as the more profitable impulse product, which is consistent with the ground-truth trajectories, once again demonstrating RL’s stronger behavioural alignment with customers movement patterns.

\begin{table}
    \centering
    \caption[Estimated impulse profit for Cluster 2 products]{
    Estimated impulse rates ($i_p$) and per-unit impulse profits ($\pi_p$) for Cluster~2 impulse products across different methods. 
    $\pi_p$ is computed using Equation~\ref{eqn:single-product-impulse-profit}, with the product price of \$3.79 (Fruits/Yogurt) and \$3.59 (Soft Drinks), and a uniform margin of 5\%. 
    Only RL identifies Soft Drinks as the more profitable impulse
product, which is consistent with the ground-truth trajectories (shown in bold).
    }
    \label{table:usecase3-estimated-profit}
    \begin{tabular}{llll}
        \toprule
        Metric & Method & Fruits/Yogurt & Soft Drinks \\
        \midrule
        \multirow{4}{*}{$i_p$} 
            & PNN    & 0.0363 & 0.013  \\
            & TSP    & 0.0363 & 0.013  \\
            & RL     & 0.0577 & 7.20   \\
            & Human & 0.115 & 3.20 \\
        \midrule
        \multirow{4}{*}{$\pi_p$} 
            & PNN    & 0.00688 & 0.00233 \\
            & TSP    & 0.00688 & 0.00233 \\
            & RL     & 0.0109  & \textbf{1.29}    \\
            & Human & 0.0218  & \textbf{0.574}  \\
        \bottomrule
    \end{tabular}
\end{table}

\boldparagraph{Step 2: Selecting the Best Shelf Location}
Intuitively, to maximize profit, the most profitable impulse product should ideally be placed on the store’s most frequently visited \textit{unoccupied} shelves.
Alternatively, one could also swap this product with highly-visited shelves that are currently \textit{occupied} by low-profit products.
While both strategies similarly increase visibility and thus purchase likelihood for the impulse product, highly-visited, low-profit products are often essential products that customers deliberately seek for.
Retailers usually place these essential products strategically to draw customers to specific areas, therefore rearranging them risks disrupting the overall traffic flow, potentially reducing visits to certain parts of the store, leading to reduced sales.
From this perspective, the first strategy—using unoccupied shelves—is less risky and more promising, and is therefore adopted in this work.

\begin{figure}
    \centering
    \includegraphics[width=\linewidth]{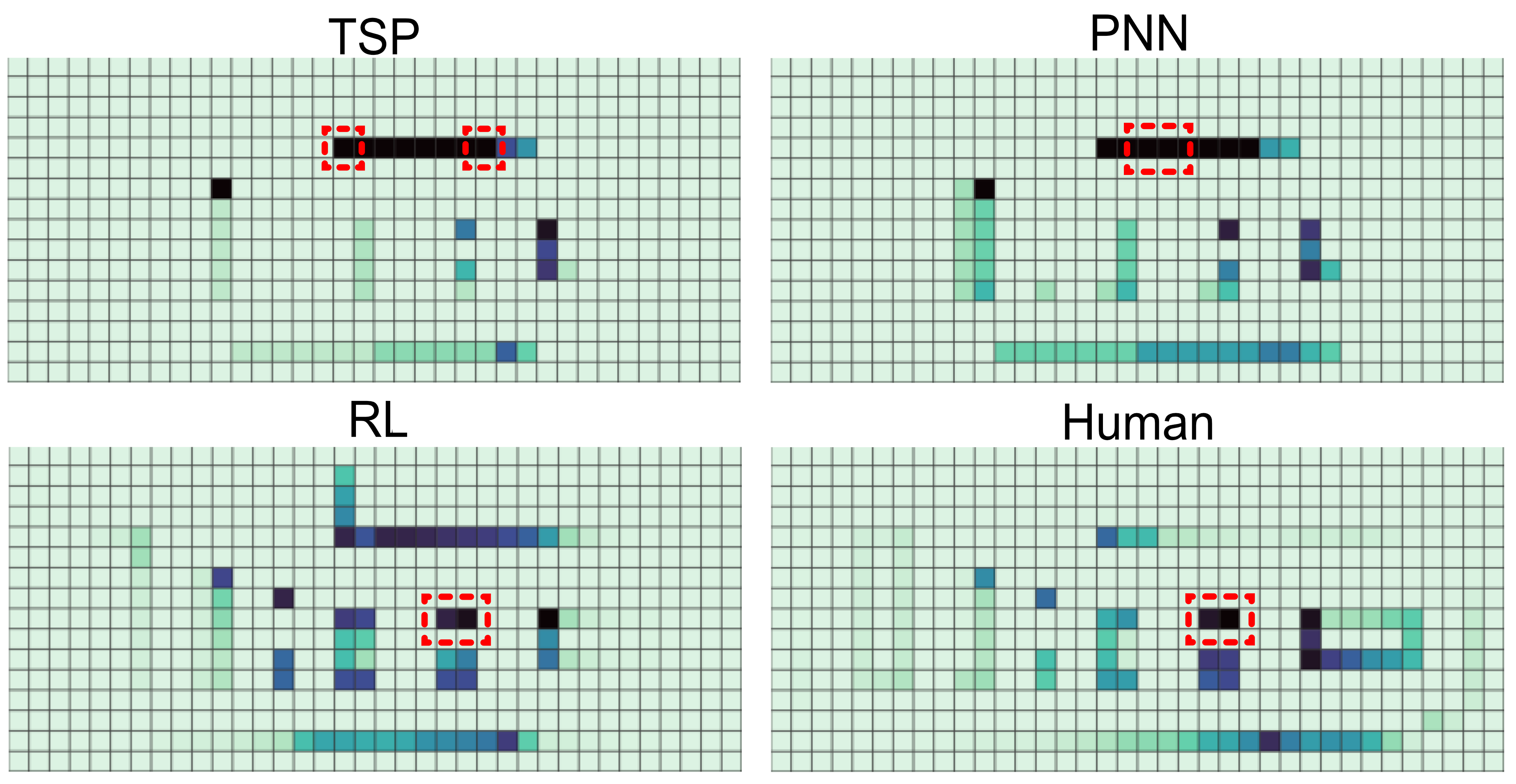}
    \caption[Shelf traffic density: top two empty shelves by visitation probability of Cluster 2 customers]{
Shelf traffic density heatmaps for Cluster 2 trajectories. 
Red dotted boxes highlight the top two \emph{empty} shelves with the highest visitation probability for each method.
Notably, RL identifies the same high-traffic shelves as the human data.
}
    \label{fig:usecase3-traffic-density}
    \Description{TODO}
\end{figure}

To implement this, we rank all unoccupied shelves by their estimated visitation probability and select the top two (since both candidate impulse products occupy two shelves). 
Shelf traffic density heatmaps with top-ranked shelves highlighted are shown in Figure~\ref{fig:usecase3-traffic-density}.

\begin{figure*}
    \centering
    \includegraphics[width=\textwidth]{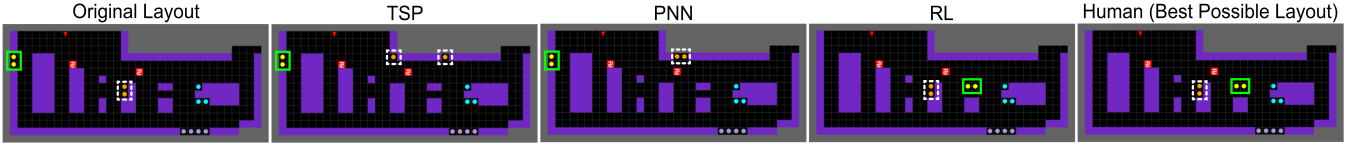}
    \caption[Layout recommendations by different methods for Cluster 2 customers]{
    Visualization of shelf layout recommendations by different methods. 
    Suggested shelf placements for Soft Drinks and Fruits/Yogurt are marked with green (solid) and white (dotted) boxes, respectively.
    Only RL replicates the ground-truth layout and achieves the highest profit gain (Table~\ref{table:final-results-profit}).
    }
    \label{fig:usecase3-new-env-layouts}
    \Description{TODO}
\end{figure*}

\boldparagraph{Step 3: Repositioning the Product}
Then, for each method, we place its chosen product on the top two unoccupied shelves.
These layout decisions can be visualized in Figure~\ref{fig:usecase3-new-env-layouts}.

\boldparagraph{Step 4: Evaluating the New Layouts}
Since only the location of impulse product changes, human trajectories that purchase essential products in the original layout remain valid, allowing them to be used for evaluation.

Actual customer trajectories from Cluster 2 are rolled-out in the modified environments, where ground-truth impulse rates (rightmost column of Table \ref{table:usecase2-impulse-rates}) are applied to determine whether a customer makes an impulse purchase upon encountering the repositioned impulse product.

Then, using the product prices and assumed profit margin, we compute the average impulse profit per customer under each method. 
As shown in Table~\ref{table:final-results-profit}, the RL-informed layout yields the highest profit gain per customer, closely matching the optimal result derived from ground-truth trajectories. 
In contrast, layouts based on TSP and PNN achieve substantially lower returns.
This is due to two key limitations: first, their predictions of most visited shelves are suboptimal. 
Second, their reliance on shortest paths prevent them from exploring other routes that customers may take when pursuing the same basket, leading to their failure in computing impulse rates for products located away from shortest paths.
By contrast, RL — under the maximum entropy framework — recovers multiple plausible customer routes, allowing it to identify the same most visited shelves and most impulsive products as those derived from ground truth human trajectories.

\begin{table}
    \centering
     \caption[Estimated impulse profit per customer before and after layout changes]{
Average impulse profit per customer (in \$) under the original and suggested layouts (visualized in Figure~\ref{fig:usecase3-new-env-layouts}), when evaluated with each method’s own trajectories or with human trajectories.
As shown in bold, the RL-informed layout yields the highest profit when evaluated with human data.
}
     \label{table:final-results-profit}
    \begin{tabular}{llll|l}
        \toprule
        Layout & TSP & PNN & RL & Human\\
        \midrule
        Original (Evaluated & 0 & 0 & 0.0140 & 0.00760 \\
        with own trajectories)&  &  &  &  \\
        Suggested (Evaluated & 0.180 & 0.171 & 0.163 & 0.162 \\
        with own trajectories)&  &  &  &  \\
        Suggested (Evaluated & 0.0900 & 0.0501 & \textbf{0.162} & \textbf{0.162} \\
        with human trajectories)&  &  &  &  \\
        \bottomrule
    \end{tabular}
\end{table}

\section{Discussion and Future Work}
In this section, we discuss key implications of our study and outline several directions for future research, organized by theme.

\paragraph{Modelling Customers with RL}
Our RL-based approach to trajectory modelling demonstrates clear behavioural advantages over heuristics, but also introduces challenges. Training robust policies is computationally expensive, whereas TSP and PNN require no training and can be computed on-the-fly. Moreover, our current RL policies must be retrained for each new layout, limiting their practicality for iterative optimization. A promising direction is to perform domain randomization on store layouts, enabling the policy to generalize across store configurations without retraining from scratch. Additionally, while the computational cost of TSP and PNN paths scale with the basket size (exponentially and quadratically, respectively), inference cost of the RL policy remains constant. Hence, for stores with large assortments, the upfront training cost of RL is amortized over repeated use.

\paragraph{Evaluation Methods}
Our evaluation showed that RL-driven estimates of shelf traffic and impulse rates can inform product repositioning decisions, yielding profit gains comparable to those obtained from ground-truth trajectories. However, this analysis was restricted to a single customer type and one impulse product, limiting generalizability. Future work could extend evaluation to multiple customer segments and product categories, and ultimately validate the approach through real-world deployments to assess its practical impact in retail environments.

\paragraph{End-to-end Layout Optimization with RL}
An intriguing direction is to extend RL beyond simulating customer trajectories to directly optimizing store layouts. Inspired by RL’s success in design tasks such as chip placement \cite{mirhoseini2021graph}, one could envision a hierarchical framework where an inner agent models customer behaviour while an outer agent iteratively adjusts product placements, enabling end-to-end layout optimization.

\paragraph{Increasing Accessibility}
Research in this area has traditionally relied on proprietary data and solving complex optimization tasks with specialized software, making replication difficult. 
Aside from reducing the need for costly real-world data through RL-generated trajectories, our work also aims to make store layout optimization more accessible. 
By releasing our code, including implementations of all methods and the environment, alongside installation/running instructions, we hope to provide a first step toward open, reproducible approaches, with the goal of enabling retailers of all sizes to benefit from these layout optimization methods.

\section{Conclusion} \label{chapter:conclusion}
Understanding customer behaviour through their movements play a critical role in store layout optimization. 
While real trajectory data offers accurate insights, it is costly to collect and often infeasible for many retailers. 
Heuristics such as the Travelling Salesman Problem (TSP) and Probabilistic Nearest Neighbour (PNN) are widely used as inexpensive substitutes, but real customer paths deviate by an average of 28\%, highlighting their limitations.

To address this gap, we framed customers as maximum entropy reinforcement learning (RL) agents, which explicitly models reward-seeking and stochasticity, better capturing bounded rationality. 
Our results show that RL-generated trajectories align more closely with customers movement than TSP or PNN, yielding more accurate predictions of shelf traffic densities and impulse purchase rates.

Beyond trajectory similarity, when used to guide a product repositioning task, RL produced layout improvements that matched those derived from ground-truth data — a result unattainable with heuristics.

Overall, RL emerges as a practical and behaviourally grounded alternative, bridging the gap between oversimplified heuristics and costly trajectory datasets, making accurate layout optimization more broadly accessible.





\newpage
\balance
\bibliographystyle{ACM-Reference-Format} 
\bibliography{references}

\end{document}